\documentclass[conference]{IEEEtran}
\IEEEoverridecommandlockouts
\usepackage{cite}
\usepackage{amsmath,amssymb,amsfonts}
\usepackage{algorithmic}
\usepackage{graphicx}
\usepackage{textcomp}
\usepackage{xcolor}
\usepackage[utf8]{inputenc}
\usepackage{subcaption}
\usepackage{lipsum}
\usepackage{float}
\usepackage{amsfonts}
\usepackage{booktabs}
\usepackage{siunitx}
\usepackage{enumitem}
\usepackage[T1]{fontenc}
\DeclareMathOperator*{\argmax}{arg\,max}

\def\BibTeX{{\rm B\kern-.05em{\sc i\kern-.025em b}\kern-.08em
    T\kern-.1667em\lower.7ex\hbox{E}\kern-.125emX}}
\begin{document}

\newlist{mylistenv}{enumerate}{3}
\newenvironment{mylist}[1]{%
    \setlist[mylistenv]{label=#1\arabic{mylistenvi}.,ref=#1\arabic{mylistenvi}}%
    \setlist[mylistenv,2]{label=#1\arabic{mylistenvi}.\arabic{mylistenvii}.,ref=#1\arabic{mylistenvi}.\arabic{mylistenvii}}%
    \setlist[mylistenv,3]{label=#1\arabic{mylistenvi}.\arabic{mylistenvii}.\arabic{mylistenviii}.,ref=#1\arabic{mylistenvi}.\arabic{mylistenvii}.\arabic{mylistenviii}}%
    \renewenvironment{mylist}{\begin{mylistenv}}{\end{mylistenv}}
    \begin{mylistenv}%
}{%
    \end{mylistenv}%
}

\title{Alignment of 3D woodblock geometrical models and 2D orthographic projection image\\
}

\author{\IEEEauthorblockN{1\textsuperscript{st} Minh Duc Nguyen}
\IEEEauthorblockA{\textit{dept. Information and Technology} \\
\textit{Vietnam National University}\\
Hanoi, Vietnam \\
}
\and
\IEEEauthorblockN{2\textsuperscript{nd} Cong Thuong Le}
\IEEEauthorblockA{\textit{dept. Information and Technology} \\
\textit{Vietnam National University}\\
Hanoi, Vietnam \\
}
\and
\IEEEauthorblockN{3\textsuperscript{rd} Trong Lam Nguyen}
\IEEEauthorblockA{\textit{dept. Information and Technology} \\
\textit{Vietnam National University}\\
Hanoi, Vietnam \\
}
}

\maketitle

\begin{abstract}
The accurate alignment of 3D woodblock geometrical models with 2D orthographic projection images presents a significant challenge in the digital preservation of Vietnam’s cultural heritage. This paper proposes a unified image processing algorithm to address this issue, enhancing the registration quality between 3D woodblock models and their 2D representations. The method includes determining the plane of the 3D character model, establishing a transformation matrix to align this plane with the 2D printed image plane, and creating a parallel-projected depth map for precise alignment. This process minimizes disocclusions and ensures that character shapes and strokes are correctly positioned. Experimental results highlight the importance of structure-based comparisons to optimize alignment for large-scale Han-Nom character datasets. The proposed approach, combining density-based and structure-based methods, demonstrates improved registration performance, offering an effective normalization scheme for digital heritage preservation.
\end{abstract}


\section{Introduction}

It is shown that recovering 3D features like depth and surface normals from a single RGB image, also known as single-view 3D is a challenging task in computer vision. Depth information is a valuable addition to computer vision applications that rely on visual (RGB) data. Time-of-flight cameras, laser range scanners, and structured light scanners are examples of traditional depth sensors. A specific single 3D objects dataset using structured light scanners includes separate data which are depth data and color data. In this case, the depth maps are not aligned perfectly with the RGB images and also have local misalignments caused by insufficient model details. The compatible depth and color gradients, and the positions of depth seeds need to be accurate.

When both color is required for neural inference to work properly and the neural data needs to be perfectly aligned with the depth data, alignment between the RGB and depth information is required. Doing semantic segmentation of color flaws and requiring to know their physical position is a classic example. Both color-based neural inference (which must be done per pixel because the network is a semantic segment) and depth information must be aligned per pixel in this situation. 


In \cite{Cadk2018AutomatedOD}, Martin Cadík et al proposed an as-rigid-as-possible deformation method that relies on the existence of sufficiently strong gradients to align mismatches between the model and photo edges in the input photograph. The misalignment of depth map edge and texture edges is one of key problems for generating high-quality synthesized views, especially for synthesizing large baseline virtual views. The misaligned depth map may cause foreground or background warped pixels to incorrect positions as inappropriate pixels are used to predict the disocclusion. To align the foreground depth edges to encompass the entire transitional color edge regions. In \cite{6738649}, a depth map preprocessing method using Watershed misalignment correction and dilation filter is presented. In a 2D sense, this method can handle sharp depth map edges lying within or outside the object borders. By using Watershed segmentation, the method can correct sharp depth map edges that are inside or outside the color image's object boundary. A foreground-biased dilation filter is used to improve the misalignment corrected depth map by assigning foreground depth values to all transitional texture edges. Besides that, Simin Zhao et al presented one type of feature in \cite{zhao2012registration}, which combines the Harris corner detector with the SIFT descriptor to effectively match 3D depth images and 2D color images and estimate the transformation homography. The features are explicitly computed at several predefined spatial scales to achieve scale invariance despite omitting the scale space analysis step of the SIFT features. Each feature point's main orientation is built in its neighborhood, and a feature descriptor is built in the direction of its main orientation. The feature space resolution is invariant due to the scale-space projection, and the feature's main orientation is rotation invariant due to the feature's main orientation. In addition, for better matching, the feature matching combines Euclidean distance with RANSAC. On the other hand, an iterative differential method introduced in \cite{8237287} uses geometric and second-order polynomial color-matching gradient-based optimization to align a colored point cloud to an image. The proposed framework introduces an algorithmic pipeline that makes use of the entire point cloud and image information to reduce the discrepancy between the point cloud colors properly projected onto the image colors. This transformation eliminates the need for additional color calibration while keeping the method simple.

The misalignment problems may occur in many problems related to reconstructing 3D woodblock characters from 2D images. Intuitively, as we have little information about the character, the first thing we want to do is to match the surface of reconstructed 3D characters with the one on the 2D papers. However, many 3D characters appear to be indifferent scales or located in abnormal positions and orientations when we transfer them to the 2D plane. In this work, we introduce a registering method that will solve such a problem. Our approach focuses mainly on finding the best match between 3D point cloud data and 2D image data. The method uses primary computer vision techniques, simple, yet powerful. 

This paper is organized as follows. In Section II.

In Section II, we present a general perspective of the woodblock character dataset. This is specific data but a general projection of a single 3D object. In Section III, we present a novel method including three phases: generating a high-resolution depth map image by identifying the projected plane of full-quality 3D point cloud data; normalizing depth-map image and 2D color image which are ingredients of a paired input; the last phase is a direction align method is used as a 2D binary character register algorithm. Experimental results and discussion are presented in Section IV. Finally, we draw our conclusions in Section V.


\section{Dataset}

\subsection{3D Dataset} 
The woodblock is a cultural heritage of Vietnam, with studies focusing on assessing the visual quality of 3D digital woodblocks for preservation purposes. Many woodblocks are lost, damaged by termites, bent, warped, cracked,... After choosing the 10 most representative woodblocks, two 3D scanners GOM ATOS and PRINCE with structured light scanning and laser scanning technology were used for woodblock scanning into the point-cloud 3D presentation. The technical requirements here require high accuracy and at the same time as soon as it is necessary to limit contact with the woodblock to avoid unwanted impact on the woodblock.



\subsection{2D Dataset}
2D data is also an annotation of 3D data which is scanned images from historical documents made of paper such as ancient books. For each specific character in point-cloud data will be mapped with a cropped image of it in 2D which is called a printed image. After the selection process, we have a total of 587 pairs of character samples with the quality criteria from medium to good from the raw data including a total of 4040 characters
\section{Proposed methods}
This section presents the steps in the 2D-3D alignment algorithm. The input 3D model is represented as a triangular mesh. The proposed algorithm can be summarized as follows: 
\begin{enumerate}
\item Determine the plane of the 3D character model, then set up the transformation matrix to transform the 3D character model so that the plane just found coincides with the $Oxy$ plane of the 2D printed image. 
\item Use parallel projection to create a depth map image of the 3D character model. 
\item Align the newly created depth-map image and the 2D-printed image. 
The above algorithm will be explained in detail along with some examples in the following sections.
\end{enumerate}

\begin{figure*}
    \centering
    \includegraphics[width=0.7\paperwidth, height=0.25\paperheight]{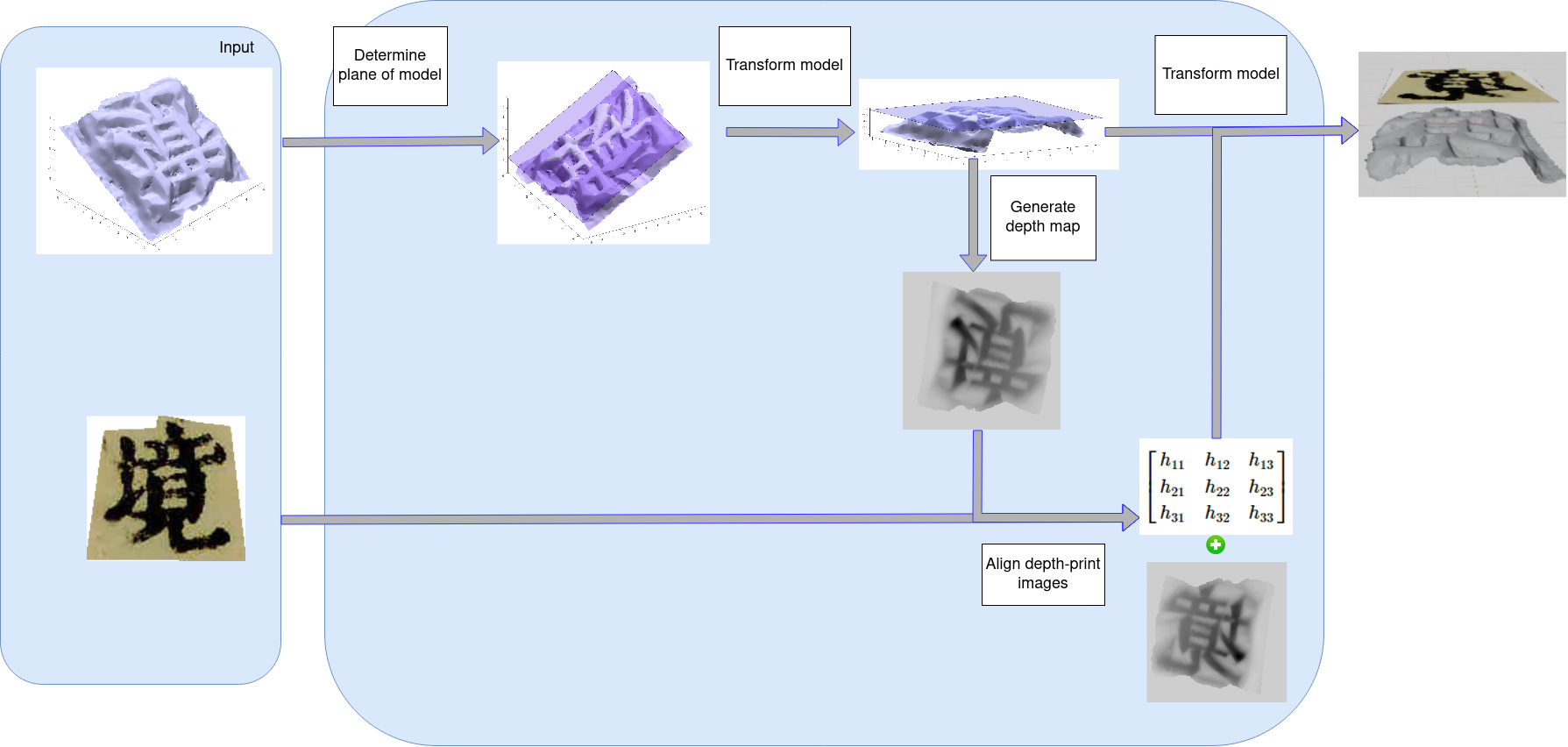}
    \caption{Overview of 2D-3D alignment pipeline}
    \label{fig:overview_pipeline}
\end{figure*}

\subsection{Determine the plane of the 3D character model}
The determination of the plane of the 3D character model is strictly based on the process of creating an ancient woodblock. The process of creating woodblocks is described in detail by Nguyen Su et al in\cite{eason1955certain}, but the article will summarize some important steps as the basis for determining the plane. At first, the artist writes the content to be engraved on the paper, then glues the surface of the paper on the flat laminated wooden board with a predetermined shape (usually a rectangular block) that is prepared in advance. As a result, the content with reverse text appears on the surface of the wooden board. Next, the artist engraves and chisels away the redundant wood parts besides the characters to obtain the characters (that are the original wood parts that are not chiseled). Thus, it is completely possible to consider the non-perforated part as a plane and the points lying on this part are the parts that are in contact with the paper when printing, this paper takes this plane as the plane of the 3D model.

\begin{figure}
  \centering
  \begin{tabular}{cc}
    \subcaptionbox{\label{img_step_1_1}}[0.4\linewidth]{\includegraphics[width=1\linewidth]{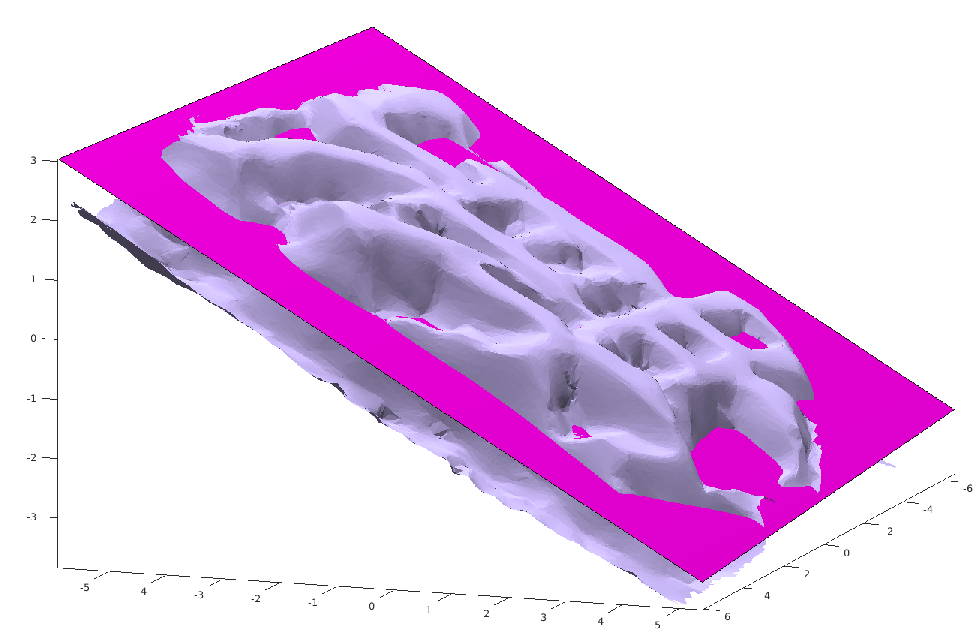}} &
    \subcaptionbox{\label{img_step_1_2}}[0.4\linewidth]{\includegraphics[width=1\linewidth]{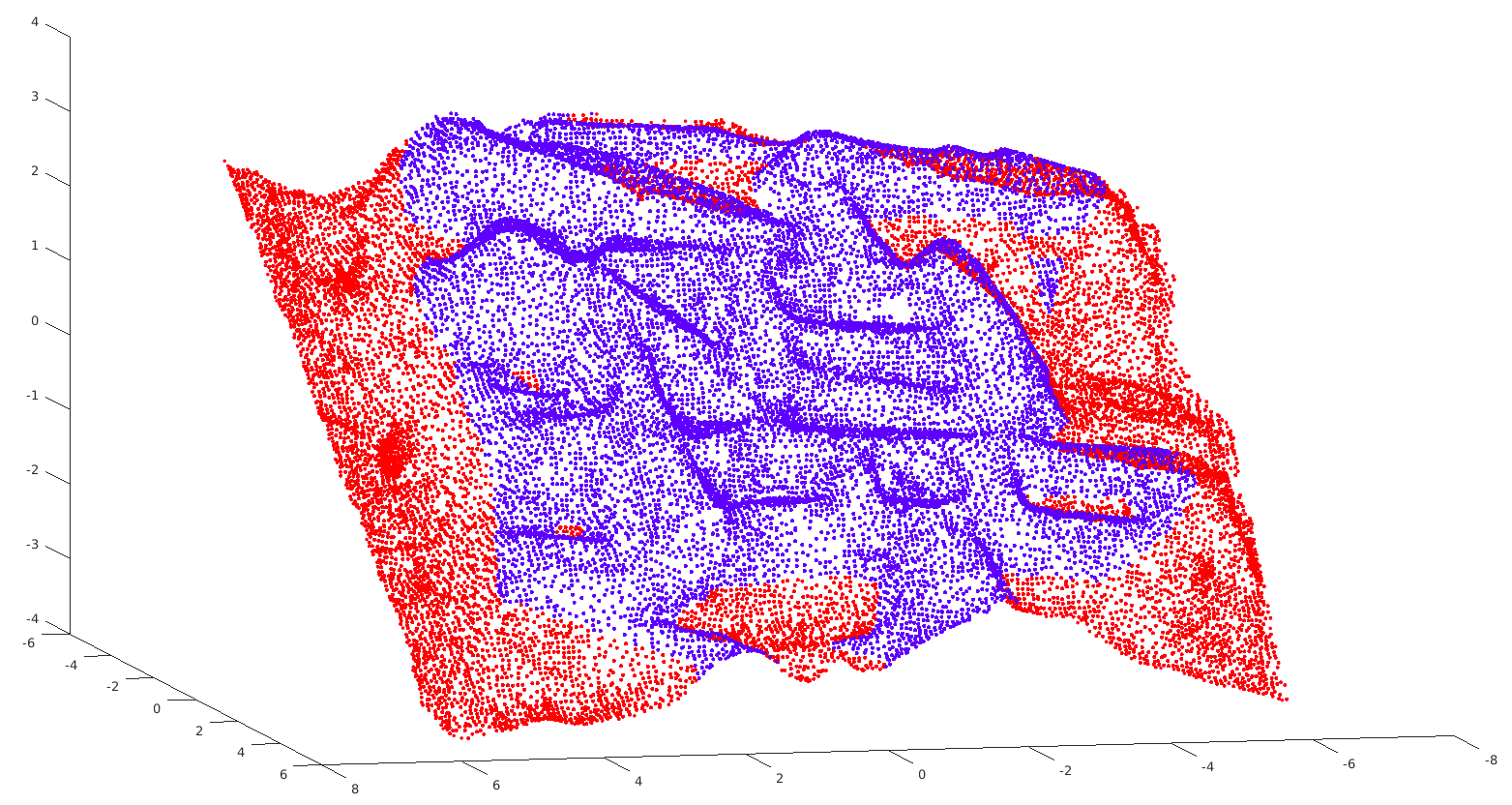}} \\
    \subcaptionbox{\label{img_step_1_5}}[0.4\linewidth]{\includegraphics[width=1\linewidth]{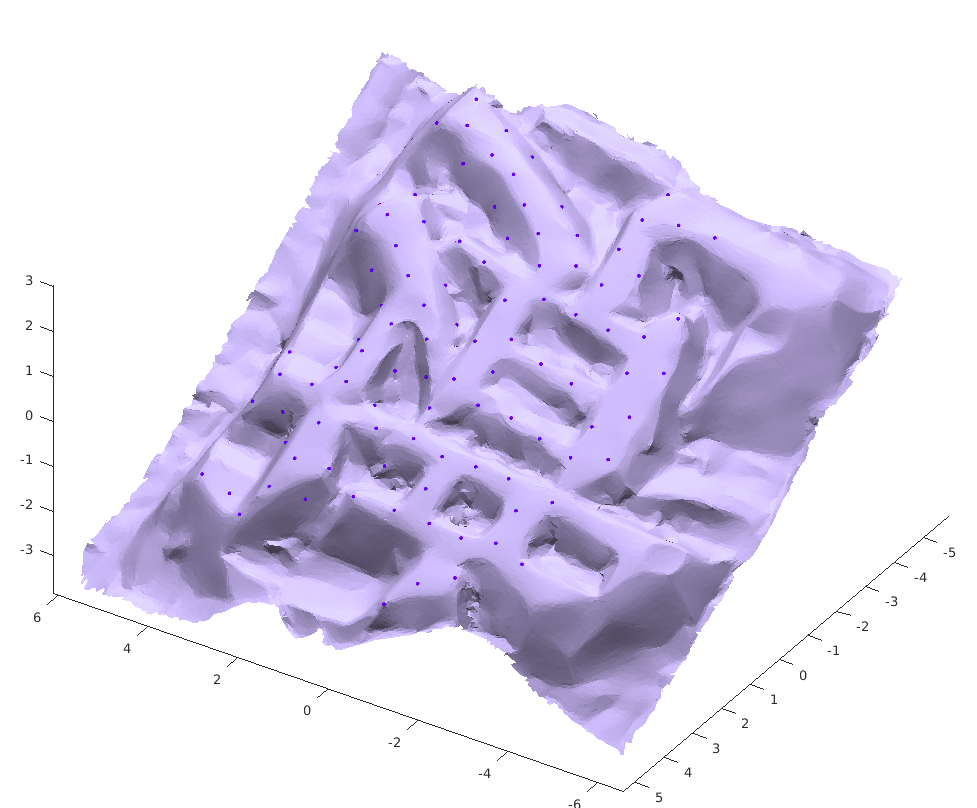}} &
    \subcaptionbox{\label{img_step_2}}[0.4\linewidth]{\includegraphics[width=1\linewidth]{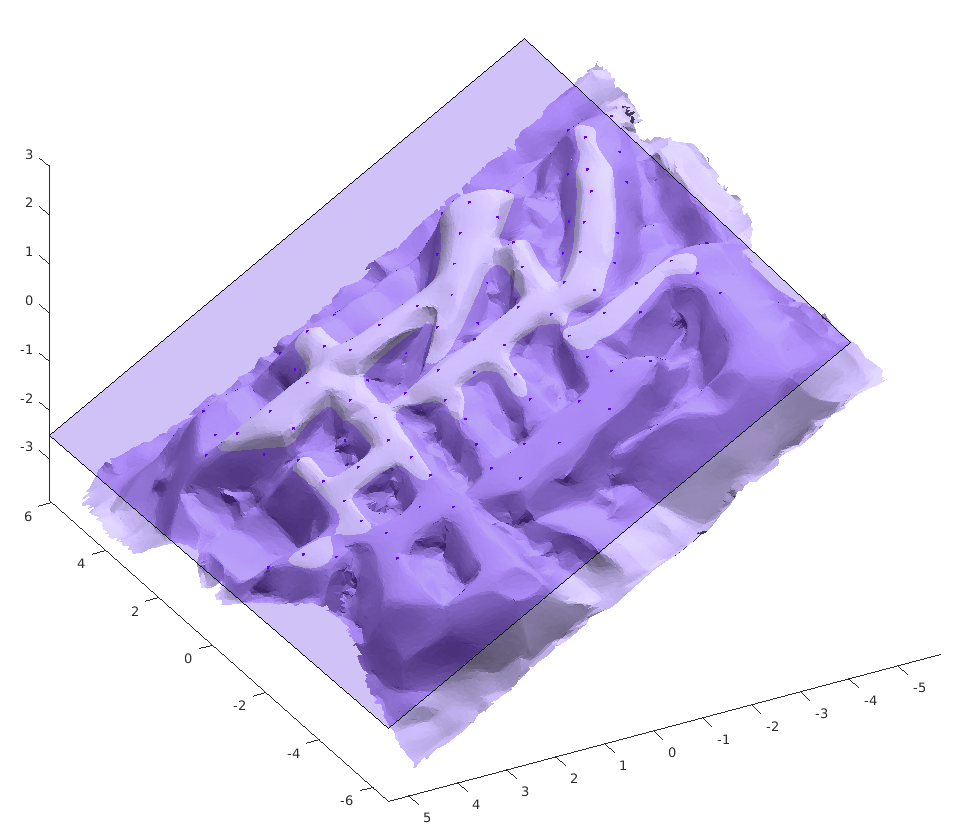}}
  \end{tabular}
  \caption{Show the results of some steps of the algorithm. (a) Result step \ref{step_1_1} (b) step \ref{step_1_2} + step \ref{step_1_3}: Bluepoint corresponds to the upper set, red corresponds to the lower point. (c) Step \ref{step_1_5} of the algorithm, the blue dots are selected points on the surface. (d) Step \ref{step_2}. \label{fig:example_steps}}
\end{figure}

The following procedure describes the procedure to be performed to determine this plane: 

Input: Woodblock character model that includes all points (called set A) related to woodblock characters such as: surface, foot, and background around the character (Illustrated as \ref{img_step_1_2}).

Output: Coefficients of plane equations in 3D space.

In order to determine the plane that approximates the model surface, the algorithm consists of two main steps: selecting points for constructing the text plane and constructing the text plane. Below is a sequence of steps of the algorithm in detail.

\begin{mylist}{S}

\item Determine the points used to construct the character plane. 

This step gets a subset of points of set A called set S, which are points lying or asymptotically lying on the plane of the character model. \label{step_1}
\begin{mylist}
\item Construct a plane based on all the vertices in set A using the PCA algorithm \cite{pearson1901liii}. \label{step_1_1} 
\item Divide all point cloud points in set A into 2 sets based on the plane built in step \ref{step_1_1} (points belonging to the same set will lie on the same side of the plane). \label{step_1_2}
\item Determine the upper set (T), the lower set (B) from the result of step \ref{step_1_2}, the set T is the set containing the points of the character surface.\label{step_1_3} To split, we perform the following steps: 
\begin{mylist}
\item We calculate the centroid of each set. 
\item For each set, average the distances from all the points to the centroid. The upper set (T) is the set with the smaller average distance. This decision is based on the fact that points in the upper part of the character have a denser and more concentrated distribution compared to the lower part. 
\end{mylist}
\item Calculate the distances of all the points of the upper set (T) to the plane constructed at B1. 
\item Determine the point lying or asymptotically lying on the target plane.

Iterate from the point with the greatest distance to the plane until there are enough points to be predetermined. The selected points need to ensure that their distance to all the pre-selected points is always greater than a given interval. \label{step_1_5}
\end{mylist}
\item Identify the coefficient of the plane equation from all points S found after at step \ref{step_1}. \label{step_2} 

Identifying the coefficient is determined by constructing a plane from the points S obtained in step \ref{step_1_5}. The PCA algorithm combined with the RANSAC algorithm \cite{fischler1981random} are used to find the plane. 
\end{mylist}

\subsection{Generate depth map from 3D model} 
\begin{figure}[h]
    \centering
    \includegraphics[width=0.9\linewidth]{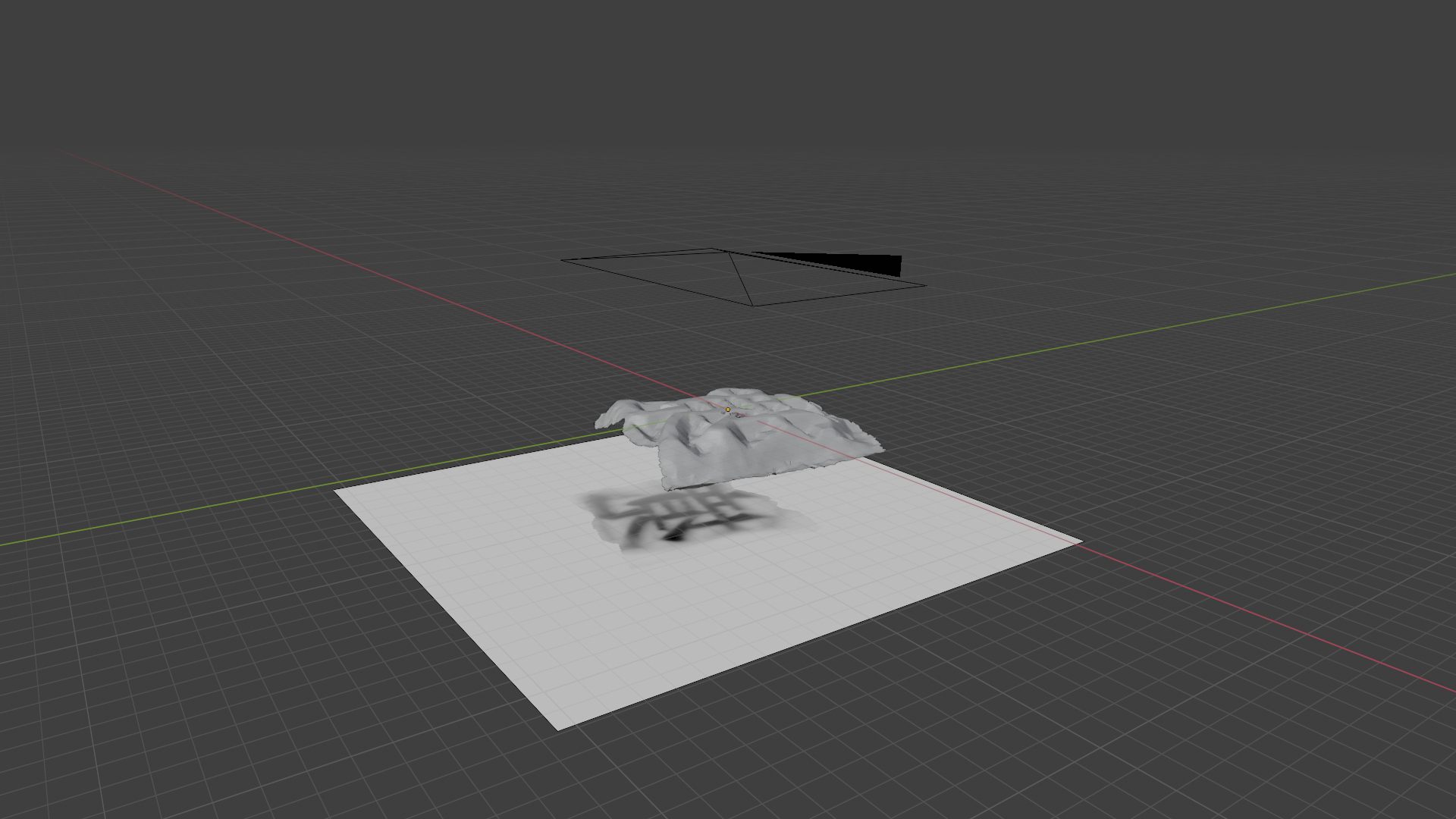}
    \caption{Parallel projection to generate depth-map image}
    \label{parallel_projection}
\end{figure}

After the coefficient of the plane equation of the 3D model is determined, we easily set up the transformation matrix to bring the newly found plane to coincide with the 0xy surface. In this step, we perform parallel projection to achieve the depth map image as shown in the figure.
Specifically, we perform parallel and inverse projection Oz, the value of each pixel is the distance between the parallel plane Oxy placed at a fixed z position and the intersection point between the projected ray and the model. (Illustrated as \ref{parallel_projection}) 

\subsection{Normalization Scheme}
\subsubsection{Binary Print Image}
Separation intensity values between character area pixels
and background are quite similar across the entire dataset. Depend on the color level, set the threshold in the range 100-127 to get the mask of the character easily.
\subsubsection{Binary Depth Image}
The depth map images are known as 2.5D images projected from top to bottom due to the 3D scanning phase. Our approach is definitely finding the best threshold for binary each specific depth map image in gray scale. A depth map image is particularly transformed to correspond to its binary of print image in both structure and human vision.

\paragraph{Structure Similarity Index metric}
The SSIM index is used for measuring the similarity between two images. The SSIM predicts image quality based on an initial uncompressed or distortion-free image as a reference.

\paragraph{Crop the wrapped character}
In order to focus on a binary object target, reducing the redundant area is the effective way. Binary image after applying this function will only keep a minimum area rectangle which includes the whole binary object.
\paragraph{Density-Based method}
This function estimates the density of the black pixels of a binary image. The density of the pixels in this region varies in the wrapped area which includes the segmented object. The density of the binary depth map needs to be the same or slightly denser than the density of the binary print image. This approach is quite good at first and can achieve an absolute approximation of the match between the binary image of the depth image and the printed image. However, this mechanism will lead to some failures though they rarely occur. Because it is not based on the structure of the object (binary characters), there will be cases where the binary image depth is not the same as the binary print image, even though they have the closest densities.
\paragraph{Structure-Based method}
Besides the method above, we propose another which is based on the core metric called Structural Similarity Index (SSIM). SSIM actually measures the perceptual difference between two similar images. It cannot judge which of the two is better so that we must determine the binary print image is the “original” and the other one has been subjected to additional processing which is our processed depth map images. The SSIM formula:
\begin{equation}
SSIM(x,y) = \frac{(2\mu_{xy} + C_1)(2\sigma_{xy} + C_2)}{(\mu_{x}^2 + \mu_{y}^2 + C_1)(\sigma_{x}^2 + \sigma_{y}^2 + C_2)}
\label{eqn:2}
\end{equation}

where $\mu$ is the measure of the average brightness of the image, $\sigma$ is the standard deviation is a measure of spread in a group of numbers. Contrast is a measure of how much the intensities are spread in an image. Very high contrast means there are very bright as well as very dark regions in an image i.e the spread of pixel intensities is high. While a low contrast indicates the pixels are in a neighborhood of similarity. $C_1$ and $C_2$ are constants to avoid instability when the denominator is close to zero. We see that this function has the quality of always being less than one and equaling one only when $\sigma_{x} = \sigma_{y}$.

Using the SSIM measure, we ignore the luminosity comparison in the original SSIM function \cite{wang2004image} as it makes no sense with binary images. The formula was simply by: 
\begin{equation}
Sim(x,y) = \frac{2\sigma_{xy} + C}{\sigma_{x}^2 + \sigma_{y}^2 + C}
\label{eqn:2}
\end{equation}
The complete calculation function:
\begin{equation}
th^* = \underset{th_{min} \leq th \leq th_{max}}{\argmax} \, \text{Sim}(W(I_1), W(I_2))
\label{eqn:2}
\end{equation}
, where $I_1$ and $I_2$ are the depth map and printed image, W is our character wrapping function and $th$ is a threshold value.
\begin{figure}[H]
	\centering
	\includegraphics[scale=0.2]{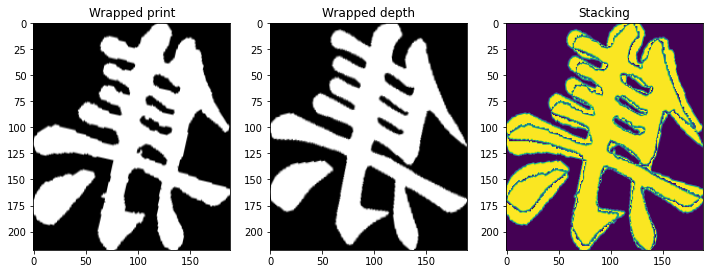}
	\caption{Example of the most appropriate threshold for depth image estimation.}
	\label{fig:celeb}
\end{figure}
The chosen binary depth image should have the best SSIM score when compared to the binary print image. This method brings an overall good result on our dataset, the SSIM metric sometimes is more reliable in our case. The reasons for some failure are most on the quality and integrity of the dataset, all of our samples were checked by human perceptual vision.

\begin{figure}[H]
  \centering
  \begin{tabular}{cc}
    \subcaptionbox{Depthmap image\label{Picture}}[0.45\linewidth]{\includegraphics[width=0.4\linewidth]{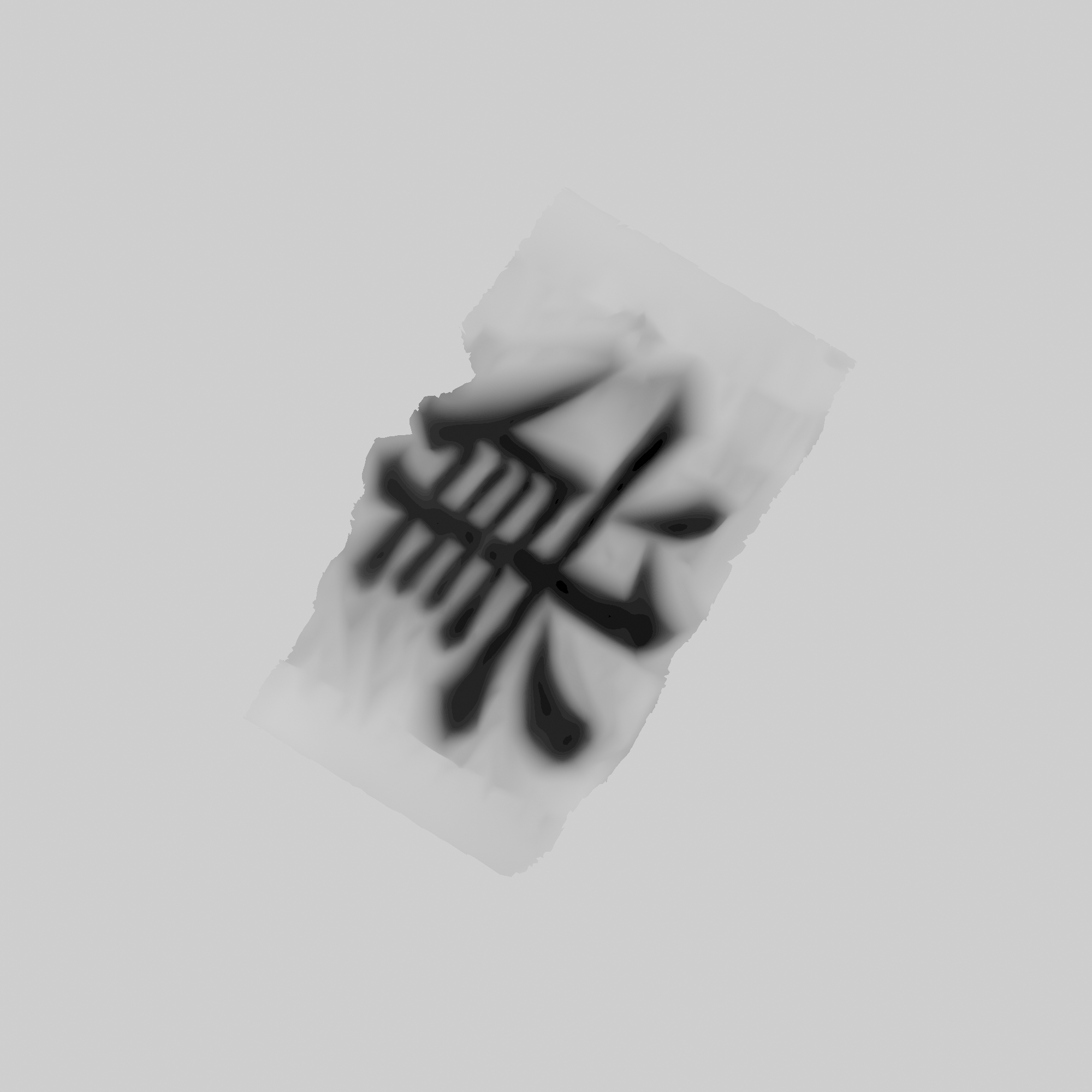}} &
    \subcaptionbox{Depth mask\label{PictureA}}[0.45\linewidth]{\includegraphics[width=0.4\linewidth]{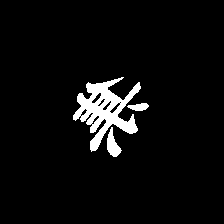}} \\
    \subcaptionbox{Print image\label{PictureB}}[0.45\linewidth]{\includegraphics[width=0.4\linewidth]{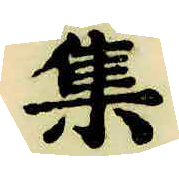}} &
    \subcaptionbox{Print mask\label{PictureC}}[0.45\linewidth]{\includegraphics[width=0.4\linewidth]{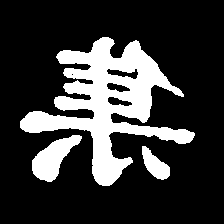}}
  \end{tabular}
  \caption{Example of final best threshold estimation.\label{fig:Four}}
\end{figure}

\paragraph{Centering character}

The image will then be shifted by a distance equal to the offset of the center of the character and the center of the image. Basically, the character will be shifted to one side with $x = x_{center of image} - x_{center of character}$ and $y = y_{center of image} - y_{center of character}$. The result we get is that the two points after the transition will coincide.
\begin{figure}[H]
  \centering
  \begin{tabular}{cc}
    \subcaptionbox{Depthmap mask.\label{Picture}}[0.45\linewidth]{\includegraphics[width=0.4\linewidth]{images/normalize_scheme/depth.png}} &
    \subcaptionbox{Depth normalized\label{PictureA}}[0.45\linewidth]{\includegraphics[width=0.4\linewidth]{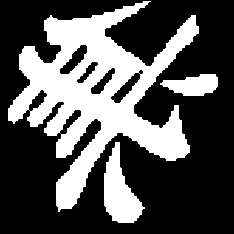}} \\
    \subcaptionbox{Print mask\label{PictureB}}[0.45\linewidth]{\includegraphics[width=0.4\linewidth]{images/normalize_scheme/print.png}} &
    \subcaptionbox{Print normalized\label{PictureC}}[0.45\linewidth]{\includegraphics[width=0.4\linewidth]{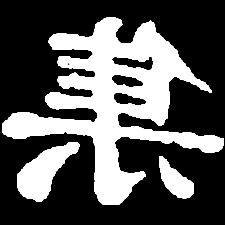}}
  \end{tabular}
  \caption{Example of centering character to the centroid.\label{fig:Four}}
\end{figure}

\subsection{Direction Align Methods}\label{SCM}
After normalization, although depth map and print images are approximately equivalent in terms of scale, those images still remain asymmetric (figure \ref{fig:Four} b,d). This problem is caused by the depth map images appearing in arbitrary places. Therefore, we address this issue by first stabilizing the print image and subsequently trying to fix the orientation and position by continuously rotating and shifting the depth map image. \\

In terms of orientation, by multiplying the depth map with a rotation matrix 
\[
\begin{bmatrix}
\cos \theta & -\sin \theta\\
\sin \theta & \cos \theta 
\end{bmatrix},
\]
where $\theta$ is a changeable angle, we get the rotated depth map images. However, for the whole dataset, finding the most suitable angle for each image usually takes a great amount of time, so we apply the divide and conquer algorithm to locate the angle faster.

\begin{figure}[H]%
    \centering
    \subfloat[\centering Angle]{{\includegraphics[width=2.5cm]{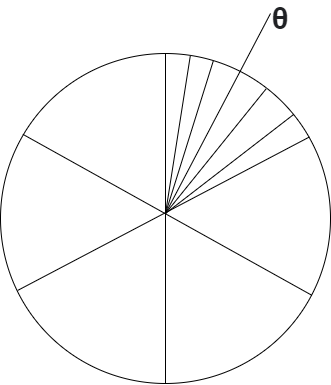}}}%
    \qquad
    \subfloat[\centering Direction padding]{{\includegraphics[width=5cm]{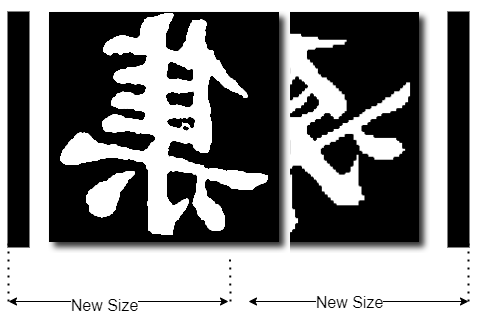}}}%
    \caption{Direction align methods}%
    \label{fig:example}%
\end{figure}

Like the name suggests, the range to find $\theta$ will be divided into 6 main sectors, whichever sectors give the highest evaluation results(IOU, SSIM, DICE) will be chosen as the following range for the upcoming iteration. Additionally, at each iteration, the characters will be “moved” by a small amount. If the characters want to move to the right, the additional padding will be added to the left.  If the characters want to move to the left, the additional padding will be added to the right and the same principles apply to top and bottom positions. Importantly, two added padding of two images must be in opposite directions. This method is to keep the size of two images stable while changing their positions and also to reduce the computational cost. 

\section{Experimental Results and Discussion}

\subsection{2D projected plane Results}
\subsection{Register Results}
This dataset is not labeled yet, hence we used our human visual system to observe all the visualized results. Following the rule: for each sample, the angle found can deviate by no more than about 5 degrees and the depth map threshold must ensure character integrity. The tests were carried out on the most accurate threshold estimation variations which are presented in the normalization scheme. The density-based technique is the first, the second one followed by the structure-based method which is applied 
Non-luminous Structure. Experiments showed that non-luminous SSIM gives only slightly better results than classical SSIM, but the computational benefits are undeniable. Composing density-based and non-luminous SSIM is the last experiment. The register results are shown in Table \ref{table:kysymys}, evaluating metrics used:
\begin{itemize}
    \item \textbf{SSIM}: The classical structure similarity index measure
    \item \textbf{IoU}: The IoU measure is used to determine how close the prediction bounding box is to the ground truth. $\mathrm{IoU}^{\mathrm{truth}}_{\mathrm{pred}}$
    \item \textbf{DICE}: The dice similarity coefficient is a reproducibility validation metric and a spatial overlap index.
\end{itemize}
\begin{table}[h]
\centering

\begin{tabular}{SSSSSSSS} \toprule
    {\textbf{Thresh metric}} & {$Acc.$} & {$SSIM$} & {$IoU$} & {$Dice$}  \\ \midrule
    {Non-luminous SSIM}  & 0.995 & 0.810 & 0.776 & 0.870  \\
    {Pixel density}  & 0.985  & 0.813 & 0.708  & 0.870   \\
    {Ensemble}  & \textbf{0.998}  & \textbf{0.821} & \textbf{0.805}  & \textbf{0.890} \\ \midrule
\end{tabular}
\caption{Register experimental results}
\label{table:kysymys}
\end{table}

\section{Conclusion}

In this paper, a unified image processing algorithm is used to review normalization schemes reported in the literature. The main purpose of this paper is to find out the best match between a 3D woodblock geometrical model and a 2D orthographic projection image. After identifying the Projected Plane, the procedure 2.5D depth map, misalignment corrected depth map method can align woodblock characters in shape and stroke, correct and place depth map image and 2D image to the right position with minimum disocclusion. Experimental results reveal that is necessary to perform a structure-comparing methodology to improve the register performance on a large-scale Han-Nom character set. Compared to density-based and structure-based methods, which solely standardize the stroke of the character image, composing two methods produces the best register performance in the normalization scheme.
\section*{Acknowledgment}

This work has been supported by Korea-Vietnam joint research project number NĐT/KR/21/01 - "AI-based Application for 3D-Model Restoration of Vietnamese and Korean Cultural Heritage". 

\bibliographystyle{plain}
\bibliography{references}

\end{document}